\documentclass{article}
\usepackage{ijcai19}

\usepackage{times}
\usepackage{soul}
\usepackage{url}
\usepackage[hidelinks]{hyperref}
\usepackage[utf8]{inputenc}
\usepackage[small]{caption}
\usepackage{graphicx}
\usepackage{amsmath}
\usepackage{booktabs}
\usepackage{algorithm}
\usepackage{algorithmic}
\usepackage{verbatim}
\usepackage{float}
\usepackage{latexsym}

\usepackage{amssymb}
\usepackage{mathtools}
\usepackage{mathrsfs}
\usepackage{verbatim}

\title{Self-Organized Inductive Reasoning with NeMuS}

\author{
Leonardo Barreto
\and
Edjard Mota
\affiliations
Institute of Computing, Federal University of Amazonas, Manaus - Brazil\\
\{lapb, edjard\}@icomp.ufam.edu.br}

\newcommand{\vect}[1]{\boldsymbol{#1}}

\begin{document}

\maketitle

\begin{abstract}
Neural Multi-Space (NeMuS)  is a weighted  multi-space representation for a portion of 
first-order logic designed for use with machine learning and neural network 
methods. It was demonstrated that it can be used to perform reasoning based
on regions forming patterns of refutation and also in the process of inductive learning in ILP-like
style. Initial experiments were carried out to investigate whether a self-organizing 
approach is suitable to generate similar concept regions according to the attributes 
that form such concepts. We present the results and make an analysis of the
suitability of the method in the process  inductive learning with NeMuS.
\end{abstract}

\section{Introduction}

Neural-Symbolic (NeSy) seeks to develop effective integration between connectionist learning and symbolic reasoning, possibly taking advantage of all statistical methods
that can be applied to the features of data perceived or on the logical structure
of symbolic information
\cite{dagstuhl17}. Moreover, the NeSy Computing community 
that has sought to integrate 
the views from AI, cognitive sciences, machine learning, ANN, computational vision 
and natural language processing and point out the main lines NeSy should go to
meet Human-Like Computaing (HLC), and in particular the Human-Like AI (HLAI) initiative. Among these points addressed, two of them are of particular interest for our experiment because it falls, somehow, in both aspects.

\emph{Statistical Relational Learning} to integrate explanation and computation of 
symbolic knowledge in deep networks. Recent results have already shown that this 
is a possible path \cite{donaserafgarcezLTN}. In the same way that the statistics of
elements of a propositional (or relational) logic program can be used to induce the 
semantics of  Artificial Neural Networks (ANN) to behave like them, e.g. C-IL$^2$P 
\cite{adg-gzCILP}, self-organization can be exploited to produce meaningful maps of 
concepts in order to ease induction of hypothesis.

\emph{Explainable AI} aims to develop AI models that are intelligible to humans, 
unlike ``black box'' models, which are efficient but difficult to extract knowledge, e.g., deep (or similar) learning without symbolic interpretation. In this direction, \emph{ patterns of concepts} can be used to justify (and explain) ``shortcuts'' to generate recursive hypothesis from very large sets of relations without the need to compute the entire path to justify it.
This is critical when the background knowledge has huge amounts of data. It could
be adequately handled as regions of concepts and categories, similar to the human 
brain map organization. This will allow symbolic deduction to be performed as matching and inductive reasoning to use weights to prune the search space of candidates for hypotheses.

The \emph{Shared Neural Multi-Space} (Shared NeMuS), is a Smarandache's multispace \cite{mao:smultispace}, or the union of $n$ spaces  $A_1,...,A_n$. As such, each $A_i$ represents the space of a characteristic distinct from total space and so it is suitable to represent different concepts of logical language. Such structure resembles the self-organization of a brain-like map as it is enhanced with adjustable weights of importance.

NeMuS was proposed in \cite{nemus4NeSy16} as a hierarchy of weighted vectors of logical elements pointing to their occurrences within a set of logical formulae in their normal disjunctive form. This structure of weights was used to generate patterns of refutation so that deduction is treated as a matching problem. The knowledge base, known as background knowledge (BK), is used to compute complementary similarities among literals and form activation regions. The present work brings another proof-of-concept that NeMuS, we claim, is suitable to deal with complex learning tasks that meet at least two of human-like AI endeavour because:

\begin{itemize}
\item the \emph{self-organizing tendency of information in our brain, occurring at all levels} is easily mapped into its hierarchy. 
\item the \emph{sort of segregation of information into distinct parts} needed  to learn about objects, relations and rules composing them can be obtained through the adaptation of its (neural) network of weights to form regions of concepts (like refutation, similarities, etc.).
\end{itemize}

Thus, NeMuS is suitable to self-organization of maps or regions of importance 
on the logical structure that have a certain \emph{relevance to what one wants 
to learn} about symbolic knowledge. This can be used, for example, to choose better deduction strategies that help reduce search space. This paper makes the following contributions: it shows how patterns of similarities among (mostly relational) logical formulae can be determined; it points out how the formation of such patterns can be used, initially, as heuristics to guide the search for consistent hypothesis in inductive learning; it brings a self-organizing perspective that one may be interested to learn about a large set of relational knowledge. 

The remainder of this paper is organized as follows: section \ref{nemusStructure} gives some brief background on NeMuS main concepts and its applications on patterns of reasoning and inductive clause learning, section \ref{som_asom} presents the self-organizing models we used in our experiments, section \ref{experiment} describes the experiment setting up
and the preliminary results found, section \ref{related} we briefly contextualize this work in relation to others similar, and finally in section \ref{conclude} concludes the work presented.
\label{intro}
\section{Background}

\subsection{Fundamentals of NeMuS structure}\label{nemusStructure}

\begin{figure}[th!]
    \centering
    \includegraphics[scale=0.60]{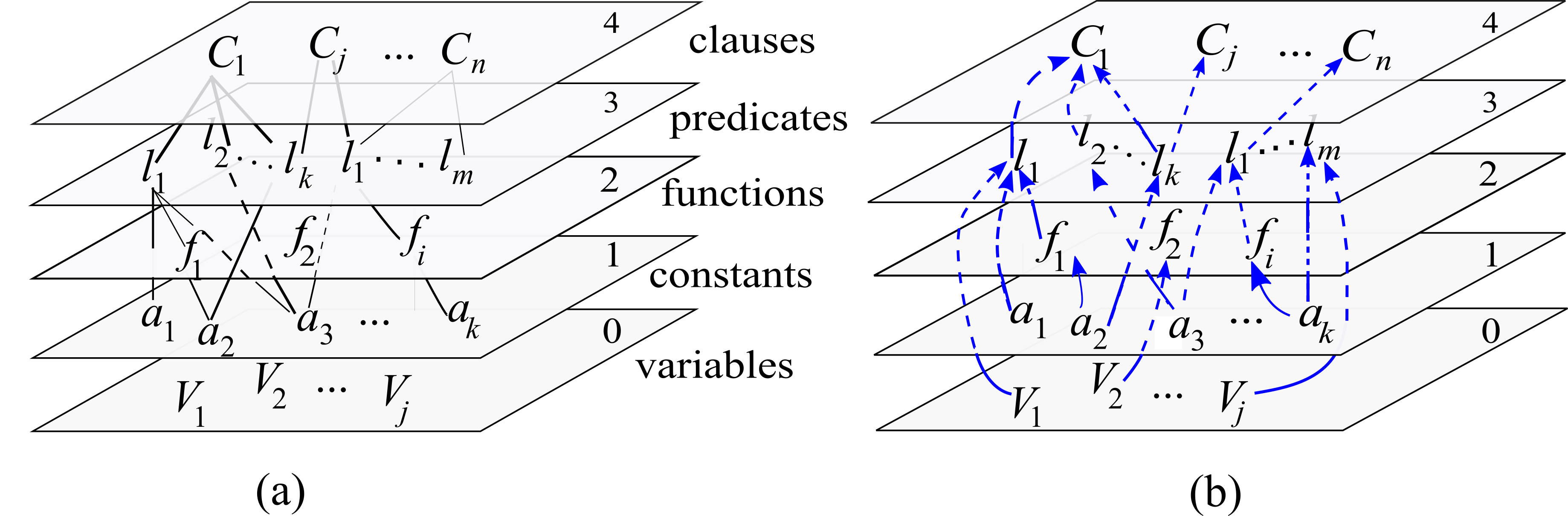}
    \caption{(a) Multi-space hierarchy, and (b) weights among  elements}
    \label{fig:forms-neural}
\end{figure}

For FOL purposes, NeMuS can be defined with  five spaces, one for each 
component of the first-order language as depicted in Figure~\ref{fig:forms-neural}.
The upward arrows illustrate the distribution of weights from the bottom up. The 
hierarchical composition of the FOL is top-down, i.e.  the semantics of a first order expression is the composition of the semantics of its subexpressions.

Note that this structure is not limited to only these spaces since clauses may influence possible worlds, possible worlds may define other higher concepts, and so on. For our purposes, we have: 0 (space) for variables, 1 for atomic constants of the Herbrand Universe, 2 for functions (suppressed here), 3 for predicates with literal instances with their own space and 4 to clauses. In what follows vectors are written $\vect{v}$, and $\vect{v}[i]$ or $\vect{v}_i$ is used to refer to an element of a vector at position $i$.

The building block of NeMuS is a 4-place vector, called T-Node, used to describe logical elements. Each element is uniquely identified by an integer code (an index) within its space. In addition, a T-Node identifies the lexicographic occurrence of the element, and (when appropriate) an attribute position. 

\newtheorem{TNode}{Definition}
\begin{TNode}[T-Node]
Let $c \in \mathbb{Z}$, $a, i, h, \in \mathbb{Z}^{+}$. A {\em T-Node (target node)\/}  is a 
quadruple $(h, c, i, a)$ that identifies an object at space $h$, with code $c$ and occurrence 
$i$, at attribute position $a$ (when it applies, otherwise 1). 
${\cal T}_{N}$ is the set of all T-Nodes.
\end{TNode}

\begin{description}
\item [NeMuS Binding] is an indexed pair $(p,w)_k$,  
        $p \in {\cal T}_{N}$, $w \in \mathbb{R}$ and $k \in \mathbb{Z}^{+}$, such that $n_{h}(p) = h$,  $n_{c}(p) = c$, $n_{a}(p) = a$ and $n_i(p) = i$. It represents the importance 
        $w$ of object $k$ over occurrence $n_i(p)$ of object $n_{c}(p)$ at space $n_{h}(p)$ in 
        position $n_{a}(p)$. 
\end{description}
\begin{description}
\item [Constant Space (1)] is a vector $\vect{C} = [ \vect{x}_1, \ldots,  \vect{x}_m ]$ , in which
         every $\vect{x}_i$ is a vector of bindings. The function $\beta$ maps a constant $i$ to 
         the vector of  its bindings $\vec{x_i}$, as above.
\end{description}

Functions, predicates and clauses are compounds forming higher spaces.
Their logical components are vectors of T-Nodes (one for each argument), and 
a vector of NeMuS bindings (simply bindings) to represent their instances.

\begin{description}
\item [Compound] in NeMuS is a vector of T-Nodes, i.e. $\vect{x}^{i}_a = [c_1, \ldots, c_m]$, 
         so that each $c_j \in {\cal T}_{N}$, and it represents an attribute  of a compound logical
         expression coded as $i$. 
\item [Instance Space] (I-Space) of a compound $i$ is the pair $(\vect{x}^{i}_{a}, \vect{w}_i)$ 
         in which $ \vect{w}_i$ is a  vector of bindings.  A vector of I-Spaces is a NeMuS    
         \emph{Compound Space (C-Space)}.
\end{description}

A literal (predicate instance), is an element of an I-Space, and so
the predicate space is simply a C-Space. Seen as compounds, clauses' attributes are the 
literals composing such clauses. 

\begin{description}
\item [Predicate Space (3)] is a pair $(\vect{C}^{+}_p, \vect{C}^{-}_p)$ in which $\vect{C}^{+}_p$ 
         and $\vect{C}^{-}_p$ are vectors of C-spaces.  
\item [Clause Space (4)]  is a vector of C-spaces such that every pair in the vector shall be  
         $(\vect{x}^{i}_{a},[ ])$. 
\end{description}

The following description is not the only way of building a NeMuS structure. For
the purpose of this work we assume no function terms, and so only 4 spaces is needed.

\newtheorem{sharedNeMuS}[TNode]{Definition}
\begin{sharedNeMuS}[Shared NeMuS]
A \emph{Shared NeMuS} for a set of coded first-order expressions is a  ordered
4-tuple(i.e. $h$ is = 4),
${\cal N}: \langle \mathcal{N}_v \mathcal{N}_c \mathcal{N}_p, \mathcal{N}_C \rangle$, in which $ \mathcal{N}_v$ is the variable space, $\mathcal{N}_c$ is the constant space, 
$\mathcal{N}_p$ is the predicate space and $\mathcal{N}_C$ is the clause space. 
\end{sharedNeMuS}

\subsection{Inductive Clause Learning (ICL) with NeMuS}

The main focus of ILP is symbolic learning of a generic logic formula $H$, called a 
hypothesis, which describes a concept, say $\alpha$, not yet defined in BK, having 
BK data and examples that characterise $\alpha$ instances or $\alpha^{+}$ positive 
examples, as well as  $\alpha^{-}$ negative examples. In other words, a formula $H$
learned is a valid hypothesis if, and only if, the union of it with BK only yields positive deductions from the concept and not the negative ones, or formally

\begin{center}
  $BK \cup \{H\} \models \alpha^{+}$, but  $BK \cup \{H\} \not \models \alpha^{-}$
\end{center}

In \cite{ilSNeMuS4NeSy17}, Mota, Howe and Garcez showed how to make use of 
NeMuS to perform the same task of inductive logic programming systems presented 
on the literature. ICL could be performed in a system called Amao, did not explicitly used the weights, but the ``intuitive use'' of them was explored to define {\em linkage patterns} between predicates of the HB and the occurrences of ground terms in positive examples. Meaningless hypotheses are pruned away as a result of {\em inductive momentum\/} between predicates connected to positive and negative examples.

The results on inductive learning in Amao show that using its shared structure leads to reliable hypothesis generation in the sense that the minimally correct ones are generated.

More recent ILP solutions proposed the use of meta-programming to define which predicates should appear in hypothesis $H$, as \cite{muggleton15meta}, like a bias in traditional ILP. However, this and other ILP approaches have to generate multiple hypothesis candidates, pair with a copy of BK, and test positive and negative deductions. A valid hypothesis search engine will be efficient if there is a partial ordering on the substitutions of hypotheses that subsume others, i.e. are more generic \cite{nienhuys1997foundations}. 

Amao use of NeMuS takes a different approach: rather than generating all possible candidates for a hypothesis, \emph{itconsiders only those whose predicates of the premises contained terms from the positive examples, while excluding from negative ones}. This is achieved by walking through the Herbrand base, from the terms of the examples, using inverse unification \cite{idestam93generalization}. 

However, when it comes to generate recursive hypothesis that demand many examples, it behaves like most approaches. This is were the importance of this experiment came about: by \emph{indicating that (relational) literals without a direct chain of attributes can be induced as recursive if the respective regions of their predicates shows that they have connections}. A parallel work to ours, \cite{motaEfficientPINemus19}, is actually making use of the results presented here (for awhile) as and heuristic to provide predicate invention on recursive hypothesis.

In order that Amao meets the desired self-organized behavior when learning (or reasoning), we took an approach that adapts itself to new information which will be very important for future neuro-symbolic applications like non-monotonic reasoning, and so on.


\section{Self-Organizing Models}\label{som_asom}
In this section, we presented a brief explanation about the Self-Organizing Maps and its variant used in this experiment, the Associative Self-Organizing Maps.

\subsection{Self-Organizing Maps}

Self Organizing Maps (SOM) is an artificial neural network which transforms a given n-dimensional pattern of data into a 1- or 2-dimensional map or grid. This transformation process is done following a topological ordering, where patterns of data (synaptic or vector weights) with similar statistical features are gathered in regions close to each other in the grid. This learning process can be classified as competitive-based because neurons compete against each other to be placed at the output layer of the neuron network, but only one wins. It is also unsupervised because the neuron network learns only with entry patterns, reorganizing itself after the first trained data and adjusting its weights as new data arrive.

A detailed description of the complete SOM algorithm is presented in \cite{kohonenBook}. In what follows, we provide a summary of the main steps of how the SOM’s learning process works.
\begin{enumerate}
    \item Initialization: at the beginning of the process, all neuron vectors have their synaptic weights randomly generated. Such vectors must have the same dimension of the entry pattern space.
    \item Sampling: a single sample $x$ is chosen from the entry pattern space and fed to the neuron grid.
    \item Competition: based on the minimum Euclidean distance criterion, the winning neuron $i(x)$ is found as follows: $$i(x) = argmin ||x - w_j||, j = 1, 2, ..., l$$ where $l$ is the number of neurons in the grid.
    \item Synaptic adaptation: after finding the winning neuron (Best-Matching Unit or BMU), all synaptic weights of each neuron vector are adjusted: $$W_j(t + 1) = W_j(t) + \eta(t) \theta_j(t)[x(t) - W_j(t)]$$ where $t$ represents the current instant, $\eta(t)$ is the learning rate which gradually decreases with time $t$, and $\theta_j(t)$ is the neighborhood function which determines the grade of learning of a neuron $j$ according to its relative distance to the winning neuron.
    \item Repeat steps 2 to 4 until no significant change happens in the topological map or achieve the number of epochs predefined.
\end{enumerate}

\subsection{Associative Self-Organizing Maps}

The A-SOM is described in \cite{johnsson2009associative} and can be considered as a SOM which learns to associate its activity with additional ancillary inputs from a number of additional SOMs. It consists of an $I \times J$ grid of neurons with a fixed number of neurons and a fixed topology. Each neuron $n_{ij}$ is associated with $r + 1$ weight vectors, where $w^{a}_{ij} \in \mathbb{R}^{n}$ is used for the main input and $w^1_{ij} \in \mathbb{R}^{m_1}$,$w^2_{ij} \in \mathbb{R}^{m_2}$, $\ldots$, $w^r_{ij} \in \mathbb{R}^{m_r}$ are used for the ancillary inputs. 

The following equations show the synaptic adaptation in the main $w^{a}_{ijk}$ and ancillary $w^{p}_{ijl}$ weight vectors where $\alpha$ is the learning rate (that decreases after each iteration), $G$ is the neighbourhood function, $c$ is the BMU index, $y^{a}_{ij}$ is the main neuron activity and $y^{p}_{ij}$ the ancillary activity. At the end of each train epoch, all the weight vectors are normalized.

\begin{equation*}
    w^{a}_{ijk}(t+1) = w^{a}_{ijk}(t) + \alpha(t)G_{ijc}(t)[x^{a}_{k}(t) -  w^{a}_{ijk}(t)]
\end{equation*}

\begin{equation*}
    w^{p}_{ijl}(t+1) = w^{p}_{ijl}(t) + \alpha(t)x^{p}_{l}(t)[y^{a}_{ij}(t) -  y^{p}_{ij}(t)]
\end{equation*}\label{backgrondInfo}
\section{Setting Up the Experiment}\label{experiment}

The experiment has two parts. The first consists of training a SOM map of concepts, $S_C$, to learn about a predicate base that describes a family genealogy using only the NeMuS constant space. The second part consists of modeling an Associative SOM, train it with the NeMuS constant and predicate spaces provided by the knowledge base (KB) and compare the two approaches. Some predicates present in the KB are described below.

\begin{table}[h]
	\begin{tabular}{ll}
		father(Jake, Bill)    & mother(Alice, Ted)   \\
		father(Jake, John)    & mother(Alice, Megan) \\
		mother(Matilda, John) & father(John, Harry)  \\
		mother(Matilda, Bill) & father(John, Susan)  \\
		father(Bill, Ted)     & mother(Mary, Harry)  \\
		father(Bill, Megan)   & mother(Mary, Susan)  \\
	\end{tabular}
	\caption{The first 12 unit clauses present in the KB.}
\end{table}

\subsection{SOM training and induction}\label{induction}

\begin{figure}[h]
	\centering
	\includegraphics[width=0.9\columnwidth]{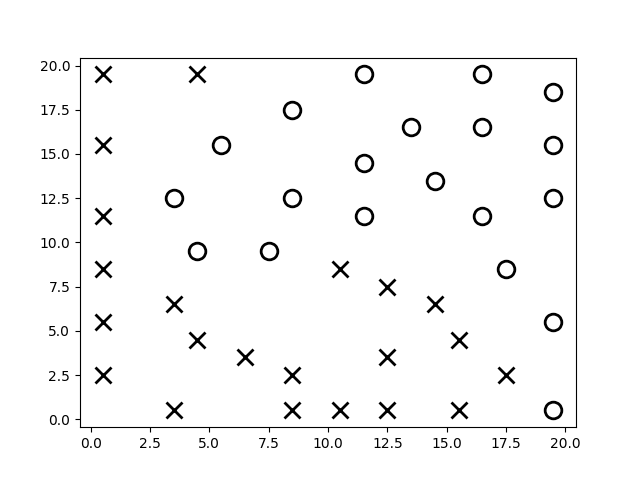}
	\caption{The 20x20 SOM grid trained using only with the NeMuS constant space.}
	\label{fig:som_before_induction}
\end{figure}

The SOM $S_C$ is generated and trained using as inputs the NeMuS constant space. The entire NeMuS structure is generated when a knowledge base is compiled. The codes for each logical element are inserted in a very efficient hashed corpus so that they are uniquely identified with space, instance, and the attribute (if it is the case) they belong to. For detailed description, we refer to \cite{nemus4NeSy16}. Our script just selects the NeMuS constant space to feed SOM training phase and it yields a map as shown in Figure \ref{fig:som_before_induction}. The circles represent the $father$ predicate instances and the $\times$`s represent the $mother$ predicate instances.

After the training, we used $S_C$ to induce rules that define predicate targets such as {\tt ancestor}. In this experiment, we used Python language and the Jupyter environment and its MiniSOM library. In the following step-by-step description of how \textit{self-organized induction} works, ${\cal N}$ is a NeMuS space instance, $E$ positive and negative examples and $V_e$ is a vector of examples.

\begin{enumerate}
	\item Normalize NeMuS constant space ${\cal N}_c$ into $\hat{{\cal N}_c}$ 
	\item Instantiate $S_C$ and train it with $\hat{{\cal N}_c}$
	\item Verify the partition of predicate regions (plot).
	\item Induction over $n$ positive or negative examples of $E$.
      \begin{enumerate}
      	\item generate NeMuS random vectors for each example $v_e \in E$
      	\item for each $v \in V_e$, do 
  	      \begin{enumerate}
  	      	\item if $v = p(a,b)$ is a positive example then:
  	      	      \begin{enumerate}
  	      	      	\item let $\beta_1 \cup \beta_2$ be a copy of  all bindings in ${\cal N}$, in which $a$  occurs as 1st argument and $b$ occurs as 2nd.
  	      	      	\item train $v_i$ with BMU weights of $\beta_1$ and $\beta_2$
  	      	      	\item select the BMU for the induction vector $v_i$
  	      	      \end{enumerate}
  	      	\item if $v = \neg q(c,d)$ is a negative example:
  	      	      \begin{enumerate}
  	      	      	\item let $\beta^{-}_1$ be from a copy of ${\cal N}$ that ignores all bindings of $c$ that it occurs as 1st argument and $\beta^{-}_2$ the ones that $d$ occurs as 2nd.
	      	      	\item train $v_i$ with BMU weights of $\beta^{-}_1 \cup \beta^{-}_2$
	      	      	\item select the BMU for the induction vector $v_i$
  	      	      \end{enumerate}
  	      \end{enumerate}
  	\item extract the new rule based on the neurons closest to the positive induction vector
  	      \begin{enumerate}
  	      	\item If $v_i$ is close to neurons representing the same predicate, then it assumes this characteristic as well. 
  	      	\item If $v_i$ and $v_j$, representing the same rule target and, at end of training, they are located in different regions, then the rule is 
  	      	      characterized by the union of these different concepts.
  	      \end{enumerate}
      	      	      	      
      \end{enumerate}
\end{enumerate}

\begin{figure}[h]
	\centering
	\includegraphics[width=0.9\columnwidth,keepaspectratio]{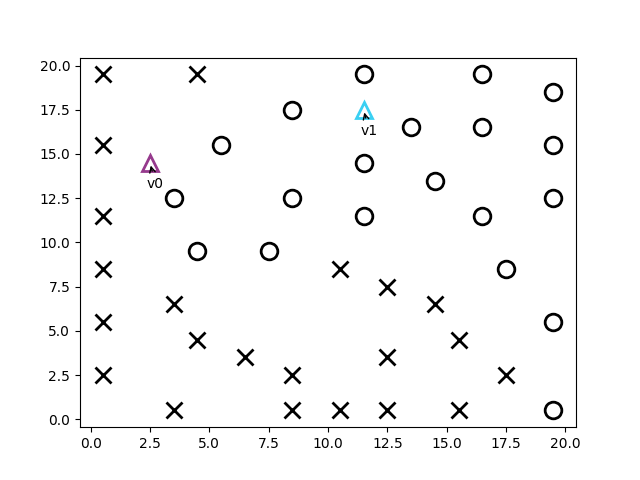}
	\caption{The 20x20 SOM grid after induce the \textit{ancestor} rule only with positive examples.}
	\label{fig:som_after_induction}
\end{figure}

For example, $ancestor (a, b)$ and $ancestor (c, d)$ are located in the mother and father regions respectively. Therefore, we can say that an ancestor can be a father or a mother. The Figure \ref{fig:som_after_induction} shows the SOM after the induction \textit{ancestor(X,Y) knowing ancestor(Jake,John) and ancestor(John,Harry)}. The triangle $v_0$ represents the induction vector of \textit{ancestor(Jake,John)} and $v_1$ represents the example \textit{ancestor(John,Harry)}. From the organization of the map, we see that both vectors are near to $father$ instances so we can assume that Jake is father of somebody that is ancestor of Harry.

\pagebreak
We have run other vector examples and the results were similar to this one, that for lack of space it is not possible to show. However, the full potential of NeMuS structure
of weights is their combination and both maps just considered the bindings of constants from the given vector examples. Next, we present an approach to do that.

\subsection{A-SOM training}

In this part, we modeled a 20x20 A-SOM called $S_A$ using as main inputs the NeMuS constant space and the NeMuS predicate space as ancillary inputs. Both spaces were generated from the knowledge base before the experiment. We choosed A-SOM to combine the different views and notations of the same base.

\begin{figure}[H]
	\centering
	\includegraphics[width=0.85\columnwidth]{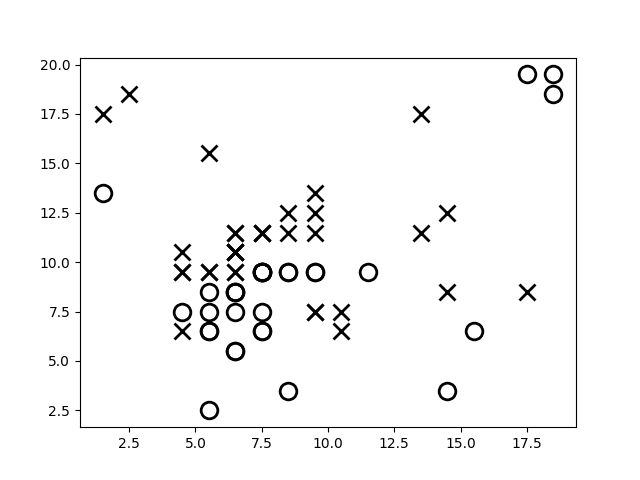}
	\caption{The 20x20 A-SOM grid trained using only with the NeMuS constant space.}
	\label{fig:asom_main}
\end{figure}

\begin{figure}[H]
	\centering
	\includegraphics[width=0.85\columnwidth]{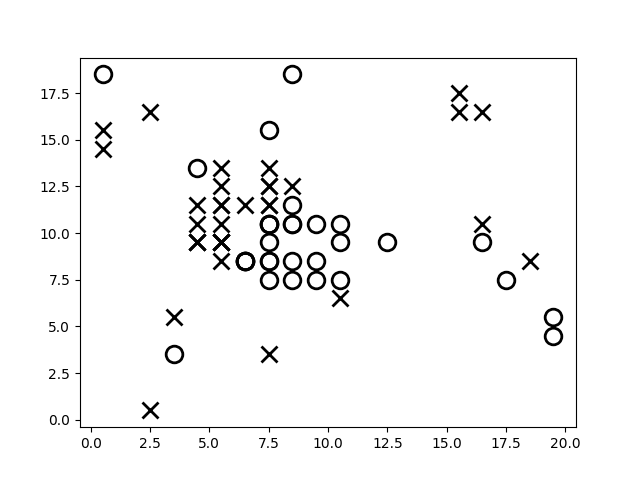}
	\caption{The 20x20 A-SOM grid trained using the NeMuS constant as main inputs and predicate spaces as ancillary inputs.}
	\label{fig:asom_main_ancillary}
\end{figure}

We trained the $S_A$ map with two approaches. The first using only the NeMus constant space as well as the first part of the experiment to compare both final results. Then, we trained using the two spaces described in the last paragraph. The comparison of these maps is showed in the Figures \ref{fig:asom_main} and \ref{fig:asom_main_ancillary}. Like the first part, the circles represent the $father$ predicate instances and the $\times$'s represent the $mother$ predicate instances.


The arrangement of regions in the associative maps was shown a little different from that presented in the SOM. This was due to the shorter amount of A-SOM training times (1,000 times) compared to SOM (10,000 times).

Because it is a more complex model, which deals with main and auxiliary inputs, it has a much larger number of vectors for synaptic adaptation (the A-SOM has $n^2(r + 1)$ associated weight vectors while the SOM has $n^2$ vectors, where $n$ is the map dimension), it would take much longer to train the two models with the same amount of iterations and we did not have enough equipment for a deeper analysis that we expect to explore in a near future.

\subsection{Dealing with Positive and negative examples}

What differentiates the training of the induction vectors from positive and negative examples is the selection of the instances that will be used for the training. If the example is positive, we select only the instances where the constants present in the example appear and respecting the order. Otherwise, we select all other instances. In the end, the positive and even negative induction vectors will be located close to some region of similarity and this is a limitation of our model: interpret the location of the induction vectors of negative examples since it can not be similar to that described section \ref{induction}.

\subsection{Results}\label{results}

We have presented and experimented a model for reasoning over a knowledge base using Self-Organizing Map (SOM) and one of its variants called the Associative Self-Organizing Map. The A-SOM develops a representation of its input space but also learns to associate its activity with the activities of an arbitrary number of ancillary inputs. In our experiments we connected an A-SOM to a list of main and ancillary inputs provided by the knowledge base using the NeMuS space notation.

As a result, we have presented a embryonic way to recognize patterns among predicates and induce rules over them using the Self-Organising Maps. However, the existence of these regions is yet to be formally proven, although it can be clearly seen in the map plot. 

\section{Related Work}\label{related}

Inductive reasoning has been the building block for the successful development of ILP approaches \cite{Muggleton2012}, as well as in the establishment of NeSy computing as an effective methodology for the integration of machine learning and reasoning \cite{2019arXiv190506088D}. Both have the benefit of having logic language as the framework to generate human-interpretable explanations, not present in other ML approaches or artificial neural network (ANN) models of learning. 

However, the recent advances in AI as a consequence of the groundbreaking achievements of Deep learning, \cite{leCunBengioDeepLearn}, with its applications, e.g. \cite{silver2017mastering} brought attention from ANN side to unveil the ``black-box'' computations, although they have surpassed human intelligent abilities in some application domains. An avalanche of works to turn deep ANN with ILP and NeSy-like features have emerged to meet the XAI challenges, e.g. \cite{dILP18}.

Our work brings a new contribution to XAI, although we have not had the time and resources to a massive data set experimentation or mathematical proofs of the existence of such regions before any computation starts. The inductive reasoning endowed with self-organized learning feature points out to a direction in which XAI system will not only be able to explain their computation, but also to give intuitive justification for reasoning with shortcuts like the one presented here and to evolve its learning and reasoning mechanisms as expected for a human-like AI system.

\section{Conclusion and Future Works}
\label{conclude}

This paper has shown some preliminary results of exploring NeMuS weighted structure to generate patterns of similarities from a small set of relational logical formulae. Those patterns can be used as a strategy to find recursive rules in a more efficient way. Although not yet integrated within Amao platform, the results presented here do support Amao learning strategy when dealing with potential recursive hypothesis. The patterns of similarities among relational logical formulae indicates how they can be used for inductive learning purposes.

For the lack of time, it was not possible to experiment the use of similar regions of concepts to guide predicate invention. Furthermore, the self-organizing approach brought interesting aspects that may be exploited to experiment like dynamic knowledge bases and non-monotonic learning and reasoning based on maps of concepts. 

Future work will focus on such aspects as well as making more efficient use of weighted structures of concepts within Amao and interact more directly with its learning components. This will help to investigate its use on learning and reasoning of complex formulae, as well as dealing with noise, uncertainty and possible worlds. We then aim to incorporate deep learning-like mechanisms for the training of massive structured datasets.
\bibliographystyle{named}
\bibliography{nesy.bib}

\begin{thebibliography}{}

\bibitem[\protect\citeauthoryear{d'Avila Garcez and
  Zaverucha}{1999}]{adg-gzCILP}
Artur d'Avila Garcez and Gerson Zaverucha.
\newblock The connectionst inductive learning and logic programming system.
\newblock {\em Applied Intelligence}, 11(1):59--77, July 1999.

\bibitem[\protect\citeauthoryear{{d'Avila Garcez} \bgroup \em et al.\egroup
  }{2019}]{2019arXiv190506088D}
Artur {d'Avila Garcez}, Marco {Gori}, Luis~C. {Lamb}, Luciano {Serafini},
  Michael {Spranger}, and Son~N. {Tran}.
\newblock {Neural-Symbolic Computing: An Effective Methodology for Principled
  Integration of Machine Learning and Reasoning}.
\newblock {\em arXiv e-prints}, page arXiv:1905.06088, May 2019.

\bibitem[\protect\citeauthoryear{Donadello \bgroup \em et al.\egroup
  }{2017}]{donaserafgarcezLTN}
Ivan Donadello, Lucianao Serafini, and Artur d'Avila Garcez.
\newblock Logic tensor networks for semantic image interpretation.
\newblock In {\em International Conference on Artificial Intelligence}, 2017.

\bibitem[\protect\citeauthoryear{Evans and Grefenstette}{2018}]{dILP18}
Richard Evans and Edward Grefenstette.
\newblock Learning explanatory rules from noisy data.
\newblock {\em Journal of Artificial Intelligence Research}, 61:1--64, January
  2018.

\bibitem[\protect\citeauthoryear{Idestam-Almquist}{1993}]{idestam93generalization}
Peter Idestam-Almquist.
\newblock Generalization under implication by recursive anti-unification.
\newblock In {\em International Conference on Machine Learning}, pages
  151--158. Morgan-Kaufmann, 1993.

\bibitem[\protect\citeauthoryear{Johnsson \bgroup \em et al.\egroup
  }{2009}]{johnsson2009associative}
Magnus Johnsson, Christian Balkenius, and Germund Hesslow.
\newblock Associative self-organizing map.
\newblock In {\em IJCCI}, pages 363--370, 2009.

\bibitem[\protect\citeauthoryear{Kohonen}{2001}]{kohonenBook}
Teuvo Kohonen.
\newblock {\em Self-Organizing Maps}.
\newblock Springer, 3rd edition, 2001.

\bibitem[\protect\citeauthoryear{Lamb \bgroup \em et al.\egroup
  }{2017}]{dagstuhl17}
Luis~C. Lamb, Tarek~R. Besolf, and Artur d'Avila Garcez.
\newblock Human-like neural symbolic computing.
\newblock Dagstuhl Reports~5, Creative Commons, May 2017.

\bibitem[\protect\citeauthoryear{Lecun \bgroup \em et al.\egroup
  }{2015}]{leCunBengioDeepLearn}
Y.~Lecun, Y.~Bengio, and G.~Hinton.
\newblock Deep learning.
\newblock {\em Nature}, 521(7553):827--832, 05 2015.

\bibitem[\protect\citeauthoryear{Mao}{2007}]{mao:smultispace}
Linfan Mao.
\newblock {An introduction to Smarandache multi-spaces and mathematical
  combinatorics}.
\newblock {\em Scientia Magna}, 3(1):54--80, 2007.

\bibitem[\protect\citeauthoryear{Mota and Diniz}{2016}]{nemus4NeSy16}
E.~de~Souza Mota and Yan~Brand{\~a}o Diniz.
\newblock {Shared Multi-Space Representation for Neural-Symbolic Reasoning}.
\newblock In Tarek~R. Besold, Luis Lamb, Luciano Serafini, and Whitney Tabor,
  editors, {\em NeSy 2016}, volume 1768. CEUR Workshop Proceedings, July 2016.

\bibitem[\protect\citeauthoryear{Mota \bgroup \em et al.\egroup
  }{2017}]{ilSNeMuS4NeSy17}
E.~de~Souza Mota, Jacob Howe, and Artur d'Avila Garcez.
\newblock Inductive learning in shared neural multi-spaces.
\newblock In Tarek~R. Besold, Artur d'Avila Garcez, and Isaac Noble, editors,
  {\em NeSy 2017}, volume 2003. CEUR Workshop Proceedings, July 2017.

\bibitem[\protect\citeauthoryear{Mota \bgroup \em et al.\egroup
  }{2019}]{motaEfficientPINemus19}
E.~de~Souza Mota, Ana Schramm, Jacob Howe, and Artur d'Avila Garcez.
\newblock Efficient predicate invention using shared nemus.
\newblock In Artur d'Avila Garcez, Freddy Lecue, and Derek Doran, editors, {\em
  NeSy 2019}. CEUR Workshop Proceedings, August 2019.

\bibitem[\protect\citeauthoryear{Muggleton \bgroup \em et al.\egroup
  }{2012}]{Muggleton2012}
Stephen Muggleton, Luc De~Raedt, David Poole, Ivan Bratko, Peter Flach, Katsumi
  Inoue, and Ashwin Srinivasan.
\newblock Ilp turns 20.
\newblock {\em Machine Learning}, 86(1):3--23, Jan 2012.

\bibitem[\protect\citeauthoryear{Muggleton \bgroup \em et al.\egroup
  }{2015}]{muggleton15meta}
S.~H. Muggleton, D.~Lin, and A.~Tamaddoni-Nezhad.
\newblock {Meta-interpretive learning of higher-order dyadic datalog: predicate
  invention revisited}.
\newblock {\em Machine Learning}, 100(1):49--73, 2015.

\bibitem[\protect\citeauthoryear{Nienhuys-Cheng and
  De~Wolf}{1997}]{nienhuys1997foundations}
Shan-Hwei Nienhuys-Cheng and Ronald De~Wolf.
\newblock {\em Foundations of Inductive Logic Programming}, volume 1228 of {\em
  Lecture Notes in Artificial Intelligence}.
\newblock Springer, 1997.

\bibitem[\protect\citeauthoryear{Silver \bgroup \em et al.\egroup
  }{2017}]{silver2017mastering}
David Silver, Julian Schrittwieser, Karen Simonyan, Ioannis Antonoglou, Aja
  Huang, Arthur Guez, Thomas Hubert, Lucas Baker, Matthew Lai, Adrian Bolton,
  Yutian Chen, Timothy Lillicrap, Fan Hui, Laurent Sifre, George van~den
  Driessche, Thore Graepel, and Demis Hassabis.
\newblock Mastering the game of go without human knowledge.
\newblock {\em Nature}, 550:354--, October 2017.

\end{thebibliography}

\end{document}